\documentclass[letterpaper, 10 pt, conference]{ieeeconf}  

\IEEEoverridecommandlockouts                              

\usepackage{graphics}
\usepackage{amsmath} 
\usepackage{amssymb}
\usepackage{cite}
\usepackage{tikz}
\usepackage{amsmath}
\usepackage{tabularx}
\usepackage{array}
\usepackage{booktabs}
\usepackage{bbm}
\usepackage{adjustbox}
\usepackage{nicematrix}
\usepackage{caption}
\usepackage{gensymb}
\usepackage{svg}
\usepackage{flushend}
\usepackage{pifont}
\usepackage{siunitx}
\usepackage{algorithm}
\usepackage{relsize}
\usepackage{subcaption}

\usepackage{tablefootnote}

\usepackage{hyperref}

\definecolor{turquoise}{cmyk}{0.65,0,0.1,0.1}
\definecolor{purple}{rgb}{0.65,0,0.65}
\definecolor{darkgreen}{rgb}{0.0, 0.5, 0.0}
\definecolor{darkred}{rgb}{0.5, 0.0, 0.0}
\definecolor{darkblue}{rgb}{0.0, 0.0, 0.5}
\definecolor{blue}{rgb}{0.0, 0.0, 1.0}
\definecolor{orange}{rgb}{1.0, 0.5, 0.0}
\definecolor{red}{rgb}{1.0, 0.0, 0.0}
\definecolor{cherry}{RGB}{186,12,47}

\title{\LARGE \bf
Fire as a Service: Augmenting Robot Simulators with\\Thermally and Visually Accurate Fire Dynamics
}

\author{Anton R. Wagner$^{1*}$,  Madhan B. Rao$^{2*}$, Helge Wrede$^{1}$, S\"oren Pirk$^{1}$, Xuesu Xiao$^{2}$
\thanks{$^{1}$ {Department of Computer Science, Kiel University, Germany}        
        {\tt\smaller \{awa, sp\}@informatik.uni-kiel.de; helge.wrede@email.uni-kiel.de}}%
\thanks{$^{2}$ Department of Computer Science, George Mason University, USA
        {\tt\smaller \{mbalajir, xiao\}@gmu.edu}}%
\thanks{$^{*}$ Equally contributing authors}%
}

\begin{document}

\makeatletter
\g@addto@macro\@maketitle{
  \begin{figure}[H]
  \setlength{\linewidth}{\textwidth}
  \setlength{\hsize}{\textwidth}
  \centering    
    \includegraphics[width=\textwidth]{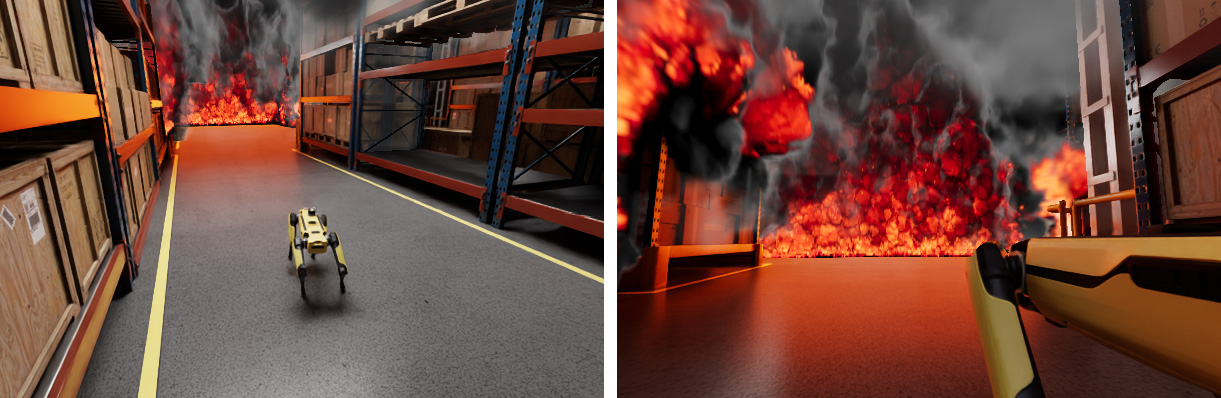}     
    \caption{Fire as a Service (FaaS) augments existing robot simulators with thermally and visually accurate fire dynamics. Here we show a Spot robot in an Isaac Sim warehouse environment affected by multiple fires from two different viewing angles.}    
    \vspace{-8mm}
    \label{fig:teaser}
  \end{figure}
}
\makeatother
\maketitle
\setcounter{figure}{1}

\begin{abstract}

Most existing robot simulators prioritize rigid-body dynamics and photorealistic rendering, but largely neglect the thermally and optically complex phenomena that characterize real-world fire environments. For robots envisioned as future firefighters, this limitation hinders both reliable capability evaluation and the generation of representative training data prior to deployment in hazardous scenarios.
To address these challenges, we introduce Fire as a Service (FaaS), a novel, asynchronous co-simulation framework that augments existing robot simulators with high-fidelity and computationally efficient fire simulations. Our pipeline enables robots to experience accurate, multi-species thermodynamic heat transfer and visually consistent volumetric smoke without disrupting high-frequency rigid-body control loops. We demonstrate that our framework can be integrated with diverse robot simulators to generate physically accurate fire behavior, benchmark thermal hazards encountered by robotic platforms, and collect realistic multimodal perceptual data. Crucially, its real-time performance supports human-in-the-loop teleoperation, enabling the successful training of reactive, multimodal policies via Behavioral Cloning. By adding fire dynamics to robot simulations, FaaS provides a scalable pathway toward safer, more reliable deployment of robots in fire scenarios.

\end{abstract}
\section{INTRODUCTION}

Robots are increasingly being envisioned for deployment in hazardous environments such as structural fires, industrial accidents, and wildfire response. In these scenarios, robots must operate under extreme heat, dynamic flame propagation, smoke-obscured visibility, and rapidly evolving thermal hazards. Developing and evaluating such systems directly in real fire environments is dangerous, expensive, and difficult to control. Simulations of fire are therefore an important tool for the development and benchmarking of robot behaviors and large-scale data collection for learning-based approaches.

Modern robot simulators provide accurate rigid-body dynamics, contact modeling, and high-quality visual rendering. However, fire is typically treated as a visual effect rather than a physical process. Flames and smoke are often rendered using particle systems or animated textures without considering physically accurate fire dynamics such as thermodynamic consistency, heat transfer modeling, or physically accurate smoke behavior. As a result, robots evaluated or trained in these environments cannot reliably assess thermal exposure, hardware survivability, and perception robustness under realistic fire conditions. For robots intended to function as future firefighters, this gap is substantial. Fire response not only demands reliable navigation and manipulation around fire, but also thermal resilience, hazard-aware planning, and perception under extreme environmental perturbations. Without thermally and visually accurate fire modeling, simulation cannot faithfully approximate the risks and operational constraints encountered in real fire scenes.

To address this gap, we introduce Fire as a Service (FaaS), a modular framework that augments existing robot simulators with accurate thermal and visual fire dynamics (Fig.~\ref{fig:teaser}). We leverage Fire-X \cite{10.1145/3763338}, an efficient state-of-the-art combustion solver, to provide accurate multi-species thermodynamics.
FaaS operates as an asynchronous, loosely coupled bridge that introduces accurate heat transfer, flame propagation, and smoke effects into standard robotic simulation. It allows for real-time robot operation and data collection by foregoing a strict synchronization between the simulations and instead favoring a non-blocking, best-effort approach in which the combustion solver always composites onto the last available pose, image and depth triplet and the robot simulator always displays the last available composite image. We define total system latency as the complete end-to-end duration: from the instant the robot simulator publishes its latest triplet to the instant the final augmented sensor array is available to the robot's perception stack.

This low latency allows FaaS to composite alpha-matted fire renderings directly onto the robot's onboard camera feeds in real-time, eliminating the need for complex reprojection or latency-masking techniques. FaaS provides four key capabilities: (1) thermally accurate hazard modeling that quantifies heat radiation and cumulative thermal risk to robotic hardware; (2) visually consistent fire and smoke dynamics that augment camera-based perception in a physically grounded manner; (3) a high-performance co-simulation architecture with sufficient temporal resolution for both human-in-the-loop teleoperation and high-frequency reactive safety controllers to train reactive, thermally-informed policies via Behavioral Cloning; and (4) engine-agnostic interoperability, with seamless integration demonstrated across Isaac Sim~\cite{isaacsim}, Gazebo~\cite{Koenig-2004-394}, and MuJoCo~\cite{todorov2012mujoco} (see examples in Fig.~\ref{fig:other-simulators}).
\section{RELATED WORK}
We review related work in existing simulation frameworks for robotics, fire and combustion, and adverse environments. 
\subsection{Robotics Simulation}
Physics-based simulators are foundational to modern robotics research, enabling scalable evaluation and sim-to-real training. Platforms such as Gazebo~\cite{Koenig-2004-394}, MuJoCo~\cite{todorov2012mujoco}, PyBullet~\cite{coumans2016pybullet} and Isaac Sim~\cite{isaacsim} provide accurate rigid-body dynamics, contact modeling, and increasingly photorealistic rendering. These systems support locomotion, manipulation, and navigation research at scale. However, despite advances in mechanical and visual realism, environmental phenomena are typically simplified. Fire, in particular, is commonly represented using particle systems or animated textures without thermodynamic consistency or heat transfer modeling. The Unreal Robotics Lab~\cite{embley2025unreal}, for example, leverages Niagara particle effects within Unreal Engine to produce visually adverse fire and smoke conditions for testing Simultaneous Localization and Mapping and visual navigation, but does not model combustion thermodynamics. As a result, existing robotic simulators cannot quantify thermal exposure, model heat-induced hardware risk, or capture physically grounded smoke–perception interactions. FaaS augments these platforms by introducing physically plausible fire dynamics while inheriting their established physics and control pipelines.

\begin{figure}
    \centering
    \includegraphics[width=\columnwidth]{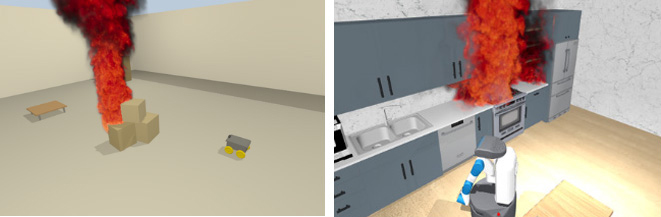} 
    \caption{FaaS superimposes scene-aware renderings of fire on renderings of established robot simulation frameworks. Besides Isaac Sim our framework supports simulators such as Gazebo (left) and MuJoCo (right).} 
    \label{fig:other-simulators}
    \vspace{-4mm}
\end{figure}

\subsection{Fire and Combustion Simulation}

Engineering-grade fire modeling tools such as NIST's Fire Dynamics Simulator (FDS)~\cite{mcgrattan2010fire} solve the full low-Mach Navier-Stokes equations with detailed combustion chemistry and have been widely validated for building safety analysis. However, the fine meshes needed to resolve turbulent flame fronts demand minutes to hours of compute per second of simulated fire, precluding interactive use~\cite{dohle2024evaluation}.

The computer graphics community has pursued the opposite trade-off. Fire and smoke have long been modeled with physically inspired fluid simulation and volumetric rendering, aiming for visually accurate results at interactive runtimes. Foundational work on stable, grid-based incompressible flow solvers enabled robust advection–projection schemes for animated gaseous phenomena \cite{10.1145/311535.311548}, and was extended to visually convincing smoke with vorticity confinement and participating-media rendering~\cite{10.1145/383259.383260}. Building on these ideas, physically based fire animation incorporated combustion-inspired source terms, temperature-driven buoyancy, and blackbody-based appearance models to produce coherent flame and smoke dynamics \cite{10.1145/566654.566643}. Subsequent methods explored production-oriented and controllable flame representations \cite{10.1145/566570.566644}, as well as higher-detail flame structure phenomena such as wrinkled and cellular flame patterns \cite{10.1145/1276377.1276436}. To meet interactive constraints, GPU-focused approaches achieved directable, high-resolution fire simulation suitable for real-time applications \cite{10.1145/1576246.1531347}. Overall, these graphics-oriented approaches emphasize controllability and visual realism, and while they often incorporate physical cues, they typically simplify or abstract detailed thermochemistry compared to Computational Fluid Dynamics techniques.

Previous work already showed how GPU-accelerated fire simulations can be integrated into third-party engines \cite{wu2023vfirelib}. Wu et al. provide a RESTful web API that allows multiple clients to connect to the simulation, which is rendered with the Unity Game Engine \cite{unity2026}. The focus of their work, however, are large-scale wildfire simulations, without using the rendered results for any downstream tasks.

FaaS builds upon the idea of connecting a fire simulation to a third-party application to get the best of both worlds. Our fire simulation is based on Fire-X~\cite{10.1145/3763338}, a GPU-accelerated hybrid solver with multi-species thermochemistry that has been validated against both real-world compartment fire experiments~\cite{mccaffrey1979} and FDS~\cite{mcgrattan2010fire}, showing strong agreement in centerline temperature profiles across heat release rates of 14--57\,kW. Crucially, it runs at rates sufficient for robot perception and planning. Contrary to the aforementioned methods, this framework also includes a rendering pipeline, which allows us to combine it with different robot simulations to produce synthetic images with high visual fidelity.

\subsection{Simulation of Adverse Environments in Robotics}
Robotics research has increasingly explored simulation of challenging environmental conditions, including terrain variability~\cite{xu2025verti}, weather effects~\cite{yang2024realistic}, sensor noise~\cite{elmquist2021sensor}, anomaly detection~\cite{wellhausen2020safe}, and domain randomization~\cite{kumar2021rma} to improve robustness and sim-to-real transfer. However, these approaches generally treat hazards as exogenous disturbances rather than physically coupled processes that affect both perception and system-level safety. 

Thermally accurate hazard modeling, such as estimating heat flux and temperature exposure to robotic platforms, remains largely absent in common robotic simulators. FaaS differs by jointly modeling thermal and visual fire dynamics and enabling quantitative benchmarking of thermal risk.

The co-simulation infrastructure required to couple a fluid solver with a robot simulator in real-time is similarly underdeveloped. Tightly coupled monolithic frameworks, such as the unified fluid-robot formulation of Lee et al.~\cite{lee2025fluidrobot}, achieve strong physical consistency but sacrifice the modularity needed to integrate with existing simulators. FaaS instead adopts an asynchronous, loosely coupled ROS~2 architecture that allows both systems to advance at independent rates.
\section{METHOD}

\begin{figure}
    \centering
    \includegraphics[width=\columnwidth]{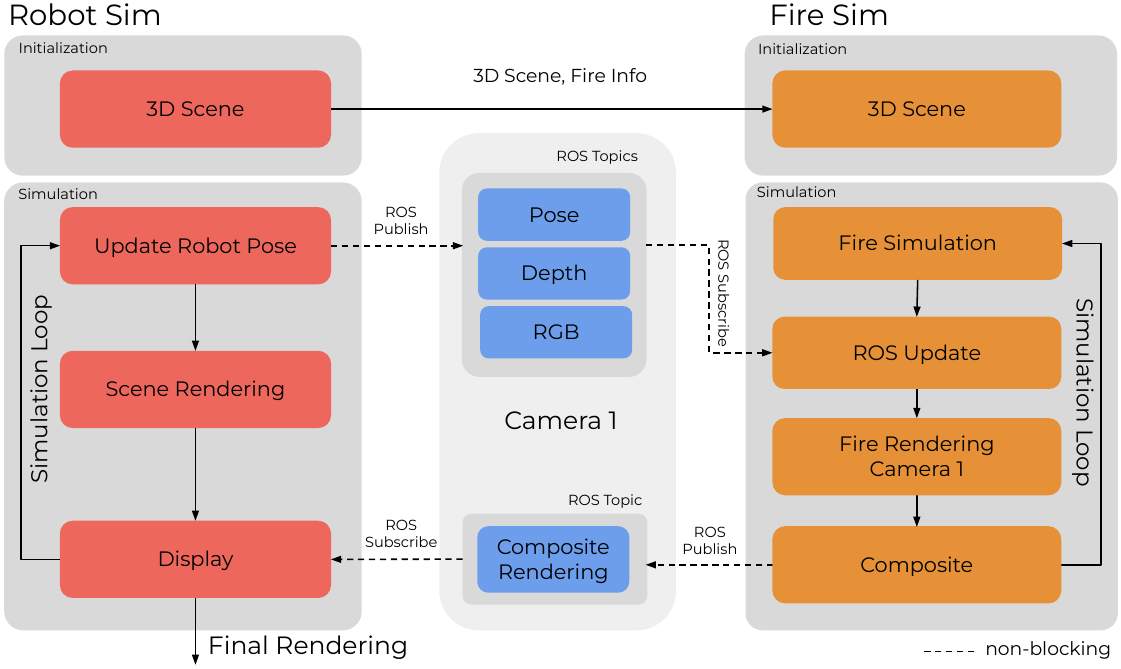}
    \caption{FaaS Overview: The system asynchronously couples a conventional robot simulator (red boxes) with an external fire simulator (orange boxes) via a non-blocking data bridge. Taking a visual camera as an example, at each timestep, the robot simulator streams camera pose and scene information to the fire simulator. For the camera (blue boxes) we establish ROS topics that we use as a means of communication between the simulators. The fire simulation rolls out thermodynamic fire dynamics and renders a viewpoint consistent with the robot's camera pose. It then generates an alpha-matted image of flames and smoke. This rendering is composited onto the robot’s RGB observations and returned, producing geometrically and temporally aligned sensor data. The loosely coupled architecture preserves high-frequency robot control while enabling physically accurate fire behavior and real-time visual augmentation.}
    \label{fig:overview}
    \vspace{-4mm}
\end{figure}

Our method is based on an asynchronous co-simulation approach that couples a conventional robot simulator with a high-fidelity fire simulator while preserving the performance and stability requirements of both systems. Instead of tightly integrating fire dynamics into the robot simulation loop, we treat the fire model as an external service that evolves independently using a physically plausible fluid and combustion solver (Fig.~\ref{fig:overview}). The key motivation of this design is to outsource complex fire dynamics to a separate simulation, while we also aim to ensure that the fire simulation remains spatially consistent with, and aware of, the 3D environment in which the robot operates, including scene geometry and obstacles that influence flame propagation and heat transfer. To achieve this, the robot simulator provides the environment representation, camera parameters, and robot pose, which are streamed to the fire simulation to maintain spatial and temporal alignment. In return, the fire simulator produces thermodynamic state information and alpha-matted renderings of flames and smoke, which are directly composited into the robot’s sensor observations. This decoupled design enables realistic, environment-aware fire behavior to be incorporated into closed-loop robotic simulation without sacrificing interactivity.

\subsection{Preliminaries}

\subsubsection{Fire Simulation}

To efficiently simulate fire we use Fire-X, the framework of Wrede et al.~\cite{10.1145/3763338}, which models combustion as a coupled, multiphase thermodynamic process across gases, liquids, and to some extent, solids. The method uses a hybrid representation that combines an Eulerian voxel grid for gaseous quantities (e.g., velocity, temperature, and species mass fractions) with Lagrangian SPH particles for liquids such as fuel or water. Combustion is modeled via a stoichiometry-aware, single-step reaction mechanism, where fuel evolves according to advection–diffusion–reaction equations, while heat is generated through an Arrhenius-type reaction rate. This heat release feeds back into the temperature field, driving buoyancy and fluid motion through a modified Navier–Stokes formulation. The simulator explicitly tracks multiple chemical species (e.g., fuel, oxygen, $CO_2$, and water vapor), enabling effects such as oxygen starvation and varying flame regimes. Additionally, phase transitions are captured through an energy-based evaporation model that couples liquid particles with the gas grid, allowing bidirectional exchange of mass, momentum, and heat. Together, these components produce physically accurate flame dynamics, including realistic plume behavior, heat transfer, and smoke formation, while remaining efficient enough for interactive simulation.

To capture thermal radiation affecting robots, we extend this framework with a particle-based radiation model. According to Merci and Beji \cite{merci2022fluid}, the thermal emission of gray bodies can be approximated with a Lambertian emission term. The radiation is discretized with particles that carry a predefined amount of energy and have a fixed velocity. The total radiative heat emission is calculated according to the Stefan-Boltzmann law. The emission pattern is based on Lambert's cosine law. The emitted particles are collected with virtual thermal sensors to capture the incident thermal radiation. The sensor readings can be used to quantify thermal damage and steer robots away from hazardous locations.

\subsubsection{Robot Simulation}
Modern robot simulators such as Isaac Sim, Gazebo, MuJoCo, and PyBullet share a common architectural foundation centered around efficient rigid-body dynamics, collision handling, and sensor simulation for closed-loop control. These platforms model robots as articulated systems with joint constraints and simulate interactions with the environment using physics engines that operate at high frequencies, typically on the order of 100 -- 400~Hz. They provide standardized abstractions for robot state (e.g., pose and velocity), environment geometry, and sensor outputs such as RGB images, depth maps, and inertial measurements. Importantly, environmental effects are generally treated as exogenous or simplified processes, with phenomena such as fire, smoke, or heat either absent or represented purely visually. This shared structure across robot simulators makes them well-suited for integration with external simulation modules, as they expose consistent interfaces for scene geometry, robot pose, and sensor data.

\subsection{Asynchronous Co-Simulation Architecture}
Our architecture connects a high-frequency robot-body engine with a fluid dynamics solver. The two simulations are decoupled and therefore it cannot be guaranteed that they update at the same rate or at the same time. A simple solution would be to strongly couple the two simulations to ensure they conform to the same update rate and delta time. This would reduce the overall simulation speed to the slowest component, making interactive evaluation and large-scale data generation impractical. To solve this frequency mismatch FaaS is designed to be asynchronous (see Fig.~\ref{fig:overview}). Camera streams from the robot simulator (RGB, Depth and Pose) are published on a best-effort basis. The fluid simulation subscribes to these topics. The fire is rendered from the camera's pose for each camera stream after each simulation step. The volumetric renderer uses the depth image to account for occlusion and composites the rendered image onto the RGB image before publishing the result back to ROS. The radiation sensor poses and their readings are exchanged using the same best-effort strategy. To account for possible discrepancies in the relative timeframes between the simulator all irradiance calculations are done using the fluid simulation's timestep. Crucially, the robot simulator is never paused. It continues executing its physics and control loop at full frequency, using the most recently received data from the fluid simulator. This represents an explicit tradeoff: without lockstep synchronization, the robot may temporarily observe stale fire data if the fluid simulator falls behind, but its control loop runs uninterrupted. The alternative, a synchronous lockstep, would guarantee temporally exact fire data but reduce the robot's effective control frequency to the fluid simulator's frame rate. FaaS currently favors real-time robot control.

\begin{figure}[t]
    \centering
    \includegraphics[width=\linewidth]{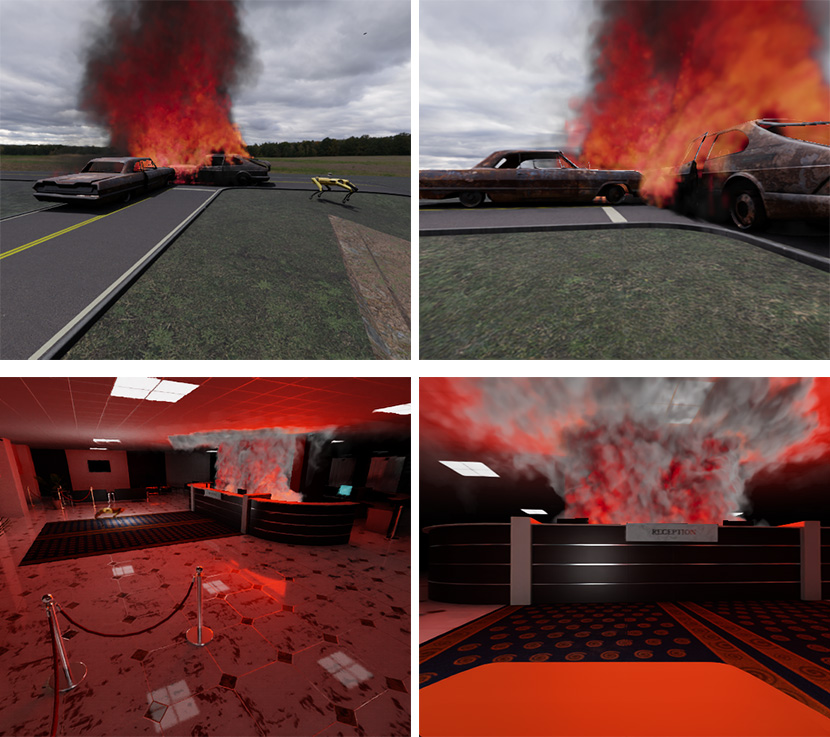}
    \caption{Two Isaac Sim scenes. Top: A two-vehicle accident scenario. A large unconfined buoyant plume demonstrates open-air advection and wind-driven flame behavior around complex vehicle geometry. The left image shows an overview of the scene while the right image shows the scene from the perspective of the robot. Bottom:  A fire plume burning behind a counter. The smoke is contained by the ceiling and starts to expand into the room. The left image shows an overview, the right image the view from the robot.}
    \label{fig:two-cars}
    \vspace{-4mm}
\end{figure}

\subsection{Virtual Thermal Sensing and Hazard Formulation}
Beyond visual rendering, FaaS translates fluid states from Fire-X into actionable robotic constraints via virtual sensors. To interface seamlessly with standard ROS~2 autonomy stacks without requiring custom planner modifications, FaaS accumulates the incident thermal radiation on a 2D plane. These thermodynamic data are normalized into discrete costmap values $[0; 100]$ and published as a standard ROS~2 \texttt{nav\_msgs/OccupancyGrid} message. This thermal costmap layer enables standard geometric path planners (e.g., Nav2) to evaluate thermodynamic gradients and autonomously avoid hazardous regions.

\begin{figure*}[t]
    \centering
    \includegraphics[width=\linewidth]{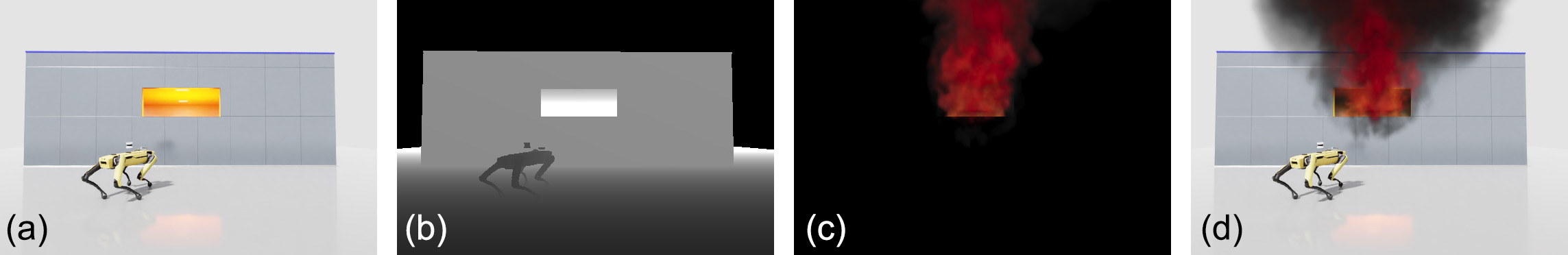}
    \caption{A fire burns behind a wall with a window. To obtain this scene we combine a rendering of a fire (c) with an RGB frame (a) and its corresponding depth map (b), which leads to a scene-aware fire rendering (d).}
    \label{fig:rgb-depth-fire-composited}
    \vspace{-4mm}
\end{figure*}

Furthermore, we implement a virtual incident thermal radiation sensor attached to the robot's body. At each simulation step, the sensor accumulates all incoming thermal radiation normalized by the sensor's surface area $\dot{q}(t)$ (\SI{}{\kilo\watt\per\square\meter}). The sensor geometry is configurable: the framework supports spherical point sensors and cuboid bodies matched to specific robot chassis dimensions. Standard path planners evaluate instantaneous costs. The energy of thermal radiation, however, also accumulates over time. This requires the framework to compute the Accumulated Thermal Dose, $D(T)$ (\SI{}{\kilo\joule\per\square\meter}), integrated over the duration $T$ (seconds) of a trajectory:
\begin{equation}
    D(T) = \int_{0}^{T} \dot{q}(t) \, dt.
\end{equation}
This cumulative quantity captures the total thermal load experienced along a trajectory, enabling comparative evaluation of path safety and hardware survivability under different planning strategies. To mitigate stochastic fluctuations from the discrete fluid solver, the raw irradiance reading is filtered through an Exponential Moving Average (EMA) with a configurable smoothing factor. Together, these capabilities enable thermally-aware planning, hardware survivability analysis, and data collection within a single unified framework.

\subsection{Visual Compositing}
Alongside thermal data, FaaS provides an optically accurate visual representation of the fire for robot RGB cameras. Our implementation achieves a mean round-trip latency of 112 ms with a standard deviation of 21 ms. The round-trip time is defined as the time it takes for the robot simulator to request and receive the thermal and visual data at a timestep. This allows the rendering pipeline and thermal data to feed directly into the robot's perception stack at interactive, closed-loop rates.

At each timestep, the FaaS bridge forwards the robot's camera pose, an RGB and a depth image to the external fluid solver. To ensure realistic fire behavior the fluid engine maintains a spatially aligned duplicate of the static environment geometry. The camera pose and depth image are used to render the fire volume with the correct perspective and occlusion. The volume rendering output is composited with the robot's RGB image based on an alpha matte, providing the perception stack with geometrically correct and temporally synchronized visual data at interactive rates. This enables the human-in-the-loop teleoperation utilized in our subsequent data collection experiments.

\section{DEMONSTRATION AND EXPERIMENTS}
We demonstrate FaaS integration into multiple robot simulators and conduct experiments to show thermally-aware path planning, learning thermal hazard avoidance, and high-frequency thermally reactive control, which are not possible with existing robot and fire simulators. All results presented in this paper were generated on a workstation equipped with an Intel i9-13900K CPU, 128 GB of RAM, and an NVIDIA RTX 4090 GPU.

\begin{figure}[t]
    \centering
    \begin{subfigure}{0.49\linewidth}
        \centering
        \includegraphics[width=\linewidth]{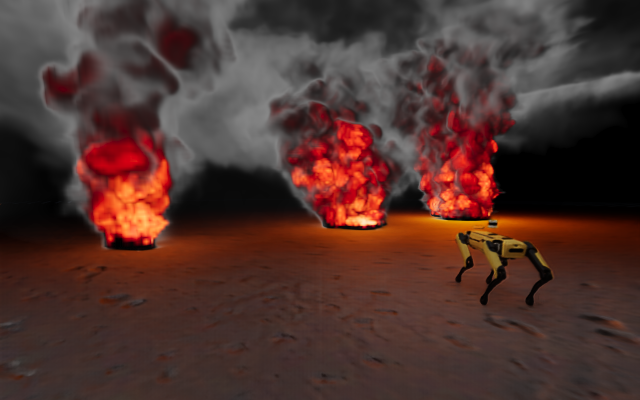}
    \end{subfigure}
    \begin{subfigure}{0.49\linewidth}
        \centering
        \includegraphics[width=\linewidth]{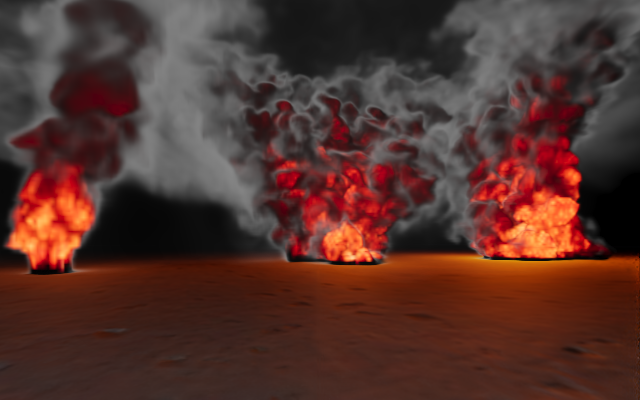}
    \end{subfigure}
    \caption{Multi-fire scenario in Isaac Sim. Three combustion sources of increasing intensity produce distinct plume heights and asymmetric smoke volumes. This environment is reused for the thermally-aware path planning experiment (Sec.~\ref{sec:path-planning}). Left: Overview. Right: Onboard camera view.}
    \label{fig:three-fires}
    \vspace{-4mm}
\vspace{4mm}
  \centering
  \includegraphics[width=\columnwidth]{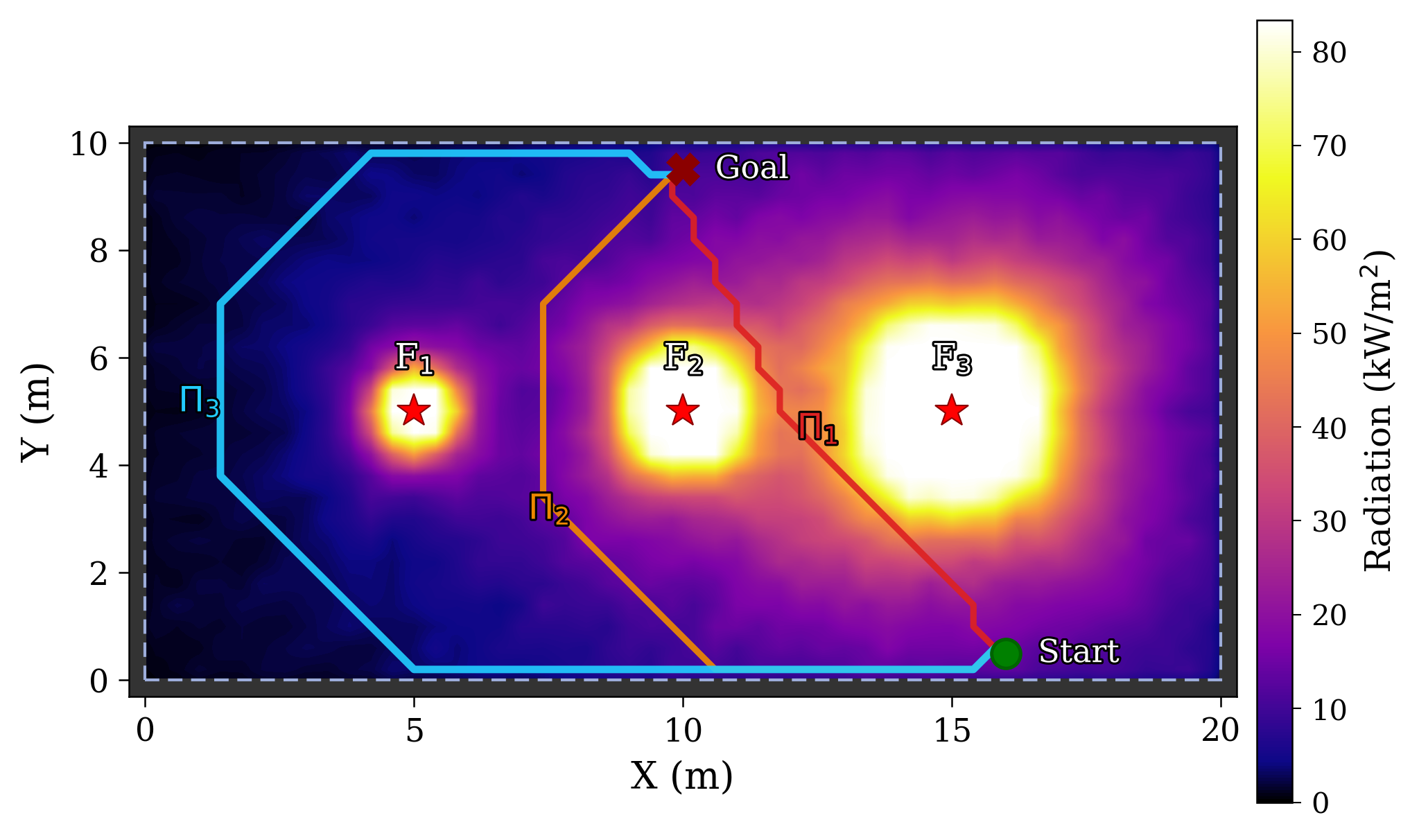}
  \vspace{-6mm}
  \caption{Time-averaged radiation costmap of the three-fire scenario with three A*-planned paths. Fire sources $F_1$ (small), $F_2$ (medium), and $F_3$ (large) are positioned along $y = 5$\,m. The heatmap shows the incident thermal radiation power and is averaged over 60~frames. Thermal doses are shown in Tab.~\ref{tab:exp2_paths}.}    
  \label{fig:thermal-paths}
  \vspace{-6mm}
\end{figure}
\begin{figure*}[t]
    \centering
    \includegraphics[width=\linewidth]{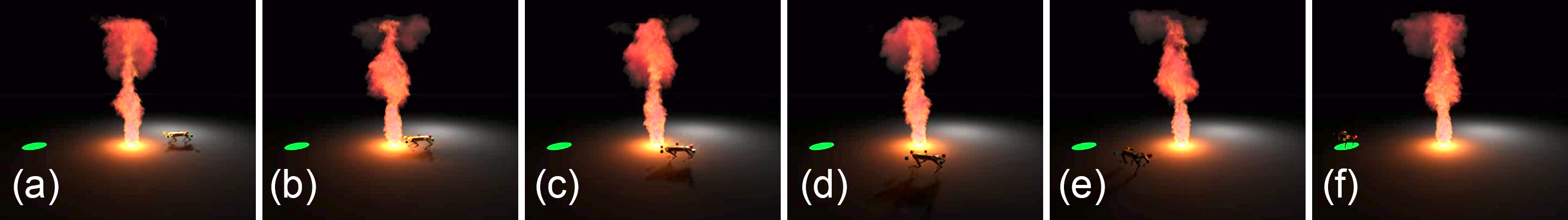}
    \caption{A temporal sequence of snapshots from the high-frequency thermally reactive control experiments. The goal position is marked in green. The robot approaches the fire from a starting position (a) trying to reach the goal behind it (b). As the incoming thermal radiation increases, the robot is steered away from the fire (c, d). The robot then proceeds to approach the goal again (e) and reaches it without further interference from the thermal irradiance (f).}
    \vspace{-4mm}
    \label{fig:reactive-control-snapshots}
\end{figure*}

\subsection{Multi-Simulator Demonstration}
\label{sec:multi-simulator-demonstration}

To validate FaaS's engine-agnostic design, we deploy the framework across three robot simulators (Isaac Sim, Gazebo, and MuJoCo), spanning indoor and outdoor fire scenarios of varying scale and confinement, as well as different robots. A warehouse scene with multiple fires between storage racks is shown in Fig.~\ref{fig:teaser}. This demonstrates the interaction between large fires and objects. In Fig.~\ref{fig:two-cars} (top) we show an outdoor two-vehicle accident where the fire produces large unconfined buoyant plumes. In confined spaces, such as a lobby (Fig.~\ref{fig:two-cars}, bottom), the fire interacts with the scene, causing the fire and the smoke to hit the ceiling of the room. Fig.~\ref{fig:rgb-depth-fire-composited} shows how the fire interacts with the environment: as the fire burns inside the room, the plume extends outside of the room through the open window. Fig.~\ref{fig:three-fires} shows a scene with three fires of different intensities. In MuJoCo, a confined kitchen stove-top fire illustrates smoke layering in a small domestic environment (Fig.~\ref{fig:other-simulators}, left). In Gazebo Harmonic, a Clearpath Jackal UGV navigates near a fire (Fig.~\ref{fig:other-simulators}, right). Across all scenes, each robot simulator only needs to subscribe to the specific ROS~2 topics to receive the fire simulation data. The image compositing works for all of the robot simulators, since the simulator renderings are also handled with topics, ensuring compatibility between the fire simulation and different robot simulations.

\subsection{Thermally-Aware Path Planning}
\label{sec:path-planning}

\paragraph{Setup} In this experiment, we position three fire sources of increasing intensity, $F_1$ (small), $F_2$ (medium), and $F_3$ (large), along the $y = 5$\,m centerline within the 20\,m $\times$ 10\,m domain (Fig.~\ref{fig:three-fires}). Each fire source is simulated with distinct properties, producing an asymmetric radiation field. A virtual sensor modeled as a cuboid matching the Spot robot chassis dimensions (1.1\,m $\times$ 0.5\,m $\times$ 0.191\,m, surface area $\approx$ 1.71\,m$^2$) \cite{spot2026specifications} must traverse from $Start = (16, 0.5)$ to $Goal = (10, 9.5)$ with walls around the simulation domain.

\begin{figure}[t]
    \centering
    \begin{subfigure}{0.8\linewidth}
        \centering
        \includegraphics[width=\linewidth]{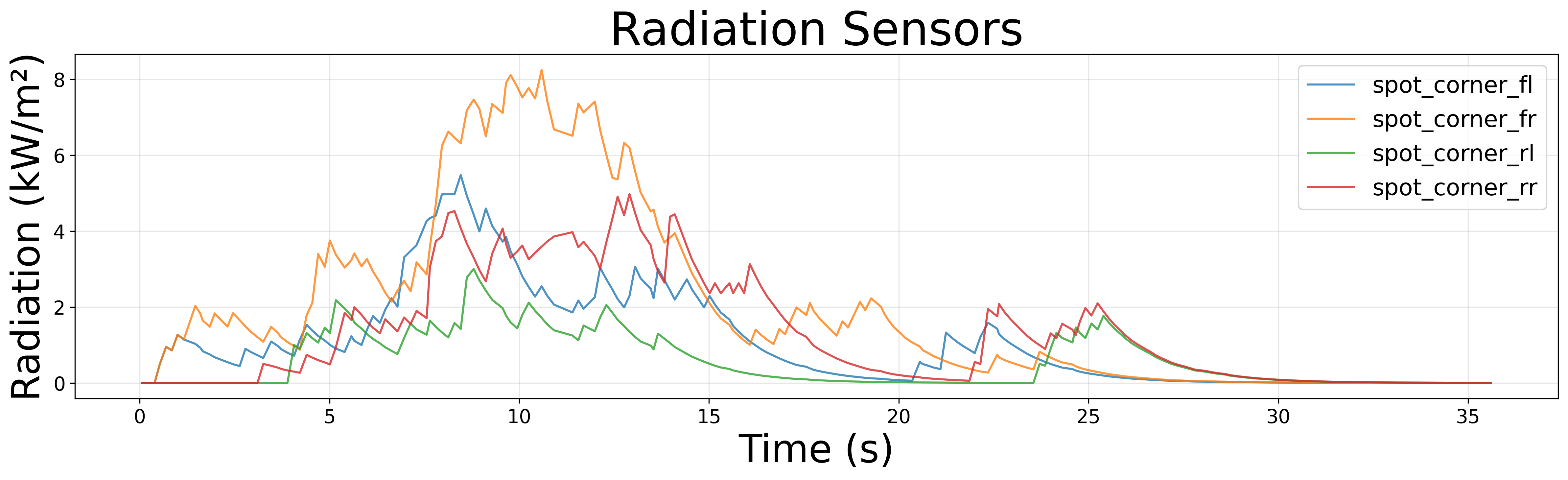}
    \end{subfigure}
    \begin{subfigure}{0.8\linewidth}
        \centering
        \includegraphics[width=\linewidth]{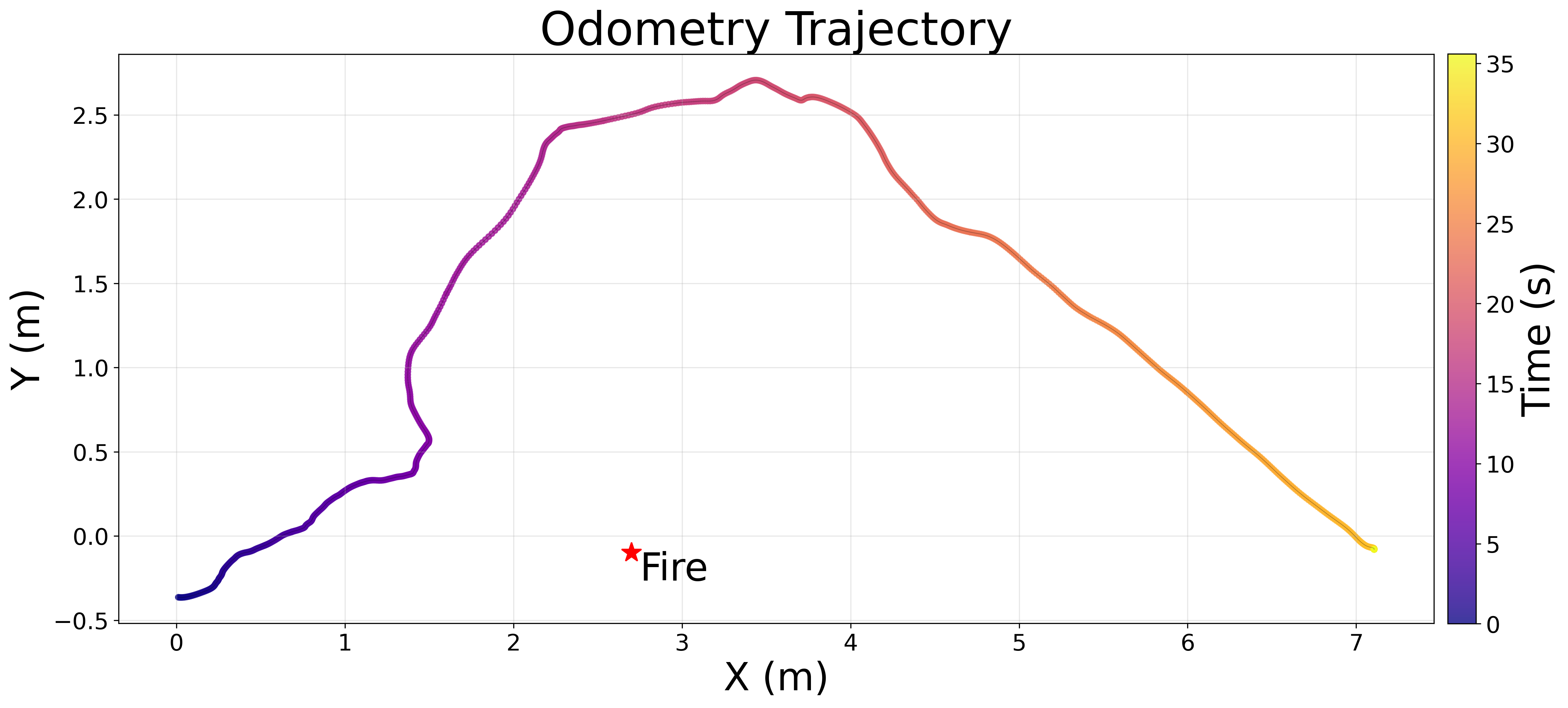}
    \end{subfigure}
    \caption{Two plots from the high-frequency thermally reactive control experiment. The top plot shows the radiation sensor readings over time. The bottom plots the corresponding odometry. As the robot gets closer to the fire, the incoming radiation increases. A peak at around 10s shows the robot nearly walking into the fire, before stopping and walking around the fire. As the robot moves away, the sensor readings decrease. Furthermore, the intensity shifts from the front to the back sensors when the robot moves next to the fire.}
    \label{fig:reactive-control-plots}
    \vspace{-4mm}
 \end{figure}

\paragraph{Radiation Costmap} We use a radiation costmap to guide the path planning. FaaS measures incoming thermal radiation onto the ground plane and publishes it as a standard ROS~2 \texttt{nav\_msgs/OccupancyGrid} message. This plane is discretized into cells with a side length of 0.4\,m. The resulting incident radiation field is linearly scaled so that the costmap values in $[0, 100]$ map to irradiance in $[0, 83]$\,\SI{}{\kilo\watt\per\square\meter}. Unlike analytical inverse-square models that assume isotropic point sources, the radiation sensors capture obstacle occlusion, plume asymmetry, and inter-source interaction (Fig.~\ref{fig:thermal-paths}). To smooth out fluctuations arising from the particle-based radiation model, we temporally average $N{=}60$ consecutive frames:
\begin{equation}
    \bar{C}[x,y] = \frac{1}{N}\sum_{i=1}^{N} C_i[x,y],
\end{equation}
where $x$ and $y$ denote the horizontal and vertical indices of the occupancy grid.

\paragraph{Path Planning} We formulate planning as A* search on the averaged costmap with an 8-connected grid. The traversal cost from cell $u$ to neighbor $v$ is:
\begin{equation}
    g(u,v) = d(u,v) \cdot \left(1 + w \cdot \frac{\bar{C}[v]}{100}\right),
\end{equation}
where $d(u,v)$ is the Euclidean distance, $\bar{C}[v]$ is the smoothed radiation value at $v$, and $w \geq 0$ controls sensitivity to radiation. Setting $w{=}0$ yields the shortest Euclidean path; increasing $w$ produces longer but thermally safer routes.

\paragraph{Dose Evaluation} To evaluate the effect of path geometry on thermal exposure without noise from controller dynamics or replanning, we evaluate each path via a deterministic sensor walk. The Spot\nobreakdash-dimensioned cuboid sensor is translated along the A* waypoints at constant speed $r = 1\,ms^{-1}$, while logging thermal radiation values returned from the fire simulation. The accumulated thermal dose is approximated by:
\begin{equation}
    D(T) \approx \sum_{k=0}^{K} \dot{q}(t_k) \cdot (t_{k+1} - t_k),
\end{equation}
where $\dot{q}(t_k)$ is the thermal radiation measured by the virtual sensor at timestep $t_k$, with simulation step $k$ and the total simulation steps $K$.

\begin{table}[t]
\centering
\caption{Thermal dose and peak irradiance measured by path following sensor.}
\label{tab:exp2_paths}
\scalebox{0.9}{
\begin{tabular}{@{}lcccc@{}}
\toprule
 & $w$ & Distance & $D(T)$ & Peak Irradiance \\
\midrule
$\sqcap_1$ & 0  & \SI{11.3}{\meter} & \SI{107.2}{\kilo\joule\per\meter\squared} & \SI{8.4}{\kilo\watt\per\meter\squared} \\
$\sqcap_2$ & 5  & \SI{16.9}{\meter} & \SI{52.0}{\kilo\joule\per\meter\squared} & \SI{3.7}{\kilo\watt\per\meter\squared} \\
$\sqcap_3$ & 30 & \SI{29.0}{\meter} & \SI{49.1}{\kilo\joule\per\meter\squared} & \SI{2.6}{\kilo\watt\per\meter\squared} \\
\bottomrule
\end{tabular}
}
\vspace{-3mm}
\end{table}

\paragraph{Results} The three cost weights produce qualitatively distinct routes (Fig.~\ref{fig:thermal-paths}). The costmap is used to predict a theoretical maximum peak irradiance which will be compared to the actual radiation sensor readings. Path $\sqcap_1$ ($w{=}0$) minimizes distance, cutting directly through the radiation field between $F_2$ and $F_3$ over \SI{11.3}{\meter} with a predicted peak irradiance of \SI{55.8}{\kilo\watt\per\meter\squared}. Path $\sqcap_2$ ($w{=}5$) detours through the inter-fire gap between $F_1$ and $F_2$, reducing the predicted peak exposure to \SI{16.7}{\kilo\watt\per\meter\squared} at \SI{16.9}{\meter}. Path $\sqcap_3$ ($w{=}30$) arcs around all three sources via the domain perimeter, achieving the lowest dose at \SI{15}{\kilo\watt\per\meter\squared} peak but requiring \SI{29.0}{\meter} to traverse. In comparison, the sensor walk (Tab.~\ref{tab:exp2_paths}) produces a general lower peak irradiance while keeping the same trends as the predicted values.

\subsection{Learned Thermal Hazard Avoidance}
\paragraph{Objective}We evaluate whether FaaS can provide sufficient thermodynamic information to train a reactive, non-visual navigation policy. Specifically, we test whether a simple learned controller can use only heat flux readings to avoid a fire hazard that it cannot visually detect.

\begin{figure}[t]
  \centering
  \includegraphics[width=\linewidth]{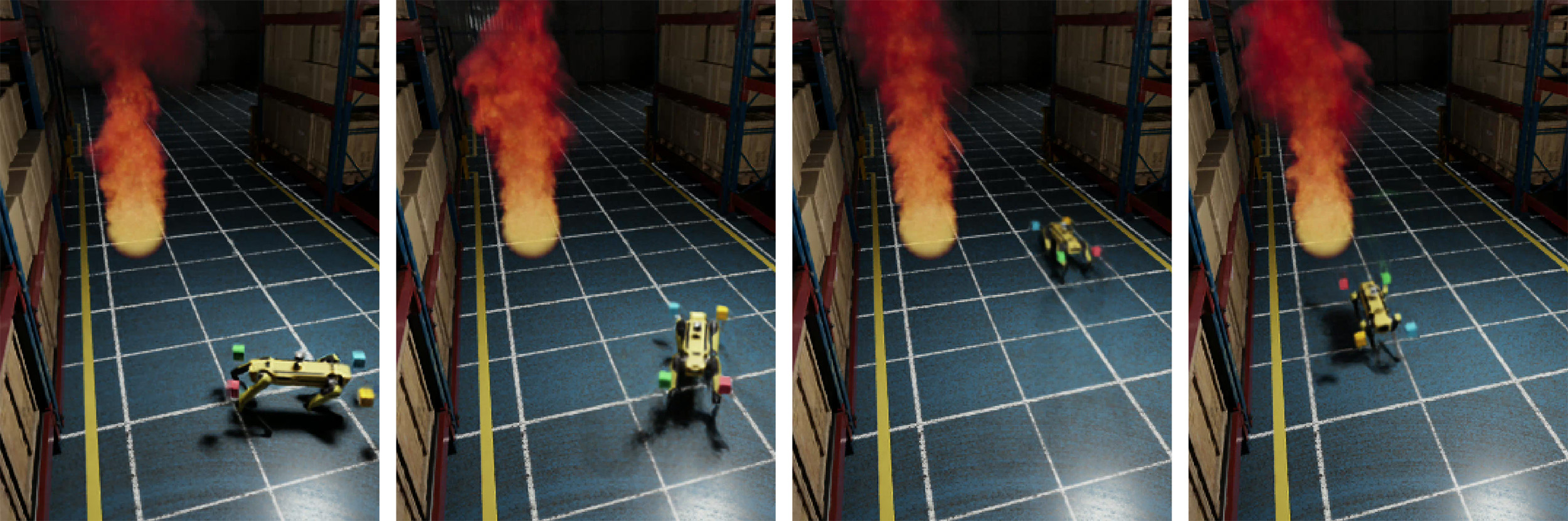}
  \caption{
  A robot equipped with four thermal sensors (colored cubes) is walking around a fire which leads to sensor readings as shown in Fig.~\ref{fig:reactive-control-plots} (top).
  }
  \label{fig:sensor_readings}
  \vspace{-6mm}
\end{figure}

\paragraph{Setup}The task consists of walking 7 meters straight forward from the starting position with a speed of \SI{1}{\meter\per\second}. \SI{2.5}{\meter} along the path a fire is either placed \SI{0.5}{\meter} to the right or the left of the straight path. To avoid excessive thermal exposure the operator, and later the learned policy, must deviate from the straight line path and later correct back to reach the original goal. The recorded data includes sensor readings from four radiation sensors placed at each corner of the robot (Fig.~\ref{fig:sensor_readings}), the robot's position and its current x/y distance to the goal. The operator was tasked with navigating from the start position to the goal while avoiding the fire visually. In total 20 paths were recorded as training data. 

\paragraph{Model}We use a simple MLP with two hidden layers of size 64 with ReLU activation. The input layer has dimension 6 for the 4 sensor readings and the delta x and y to the goal. The output layer is $\tanh$ activated and outputs a scalar representing the steering angle from -90 to 90 degrees.

\begin{figure}[t]
  \centering
  \begin{subfigure}{0.8\linewidth}
      \centering
      \includegraphics[width=\linewidth]{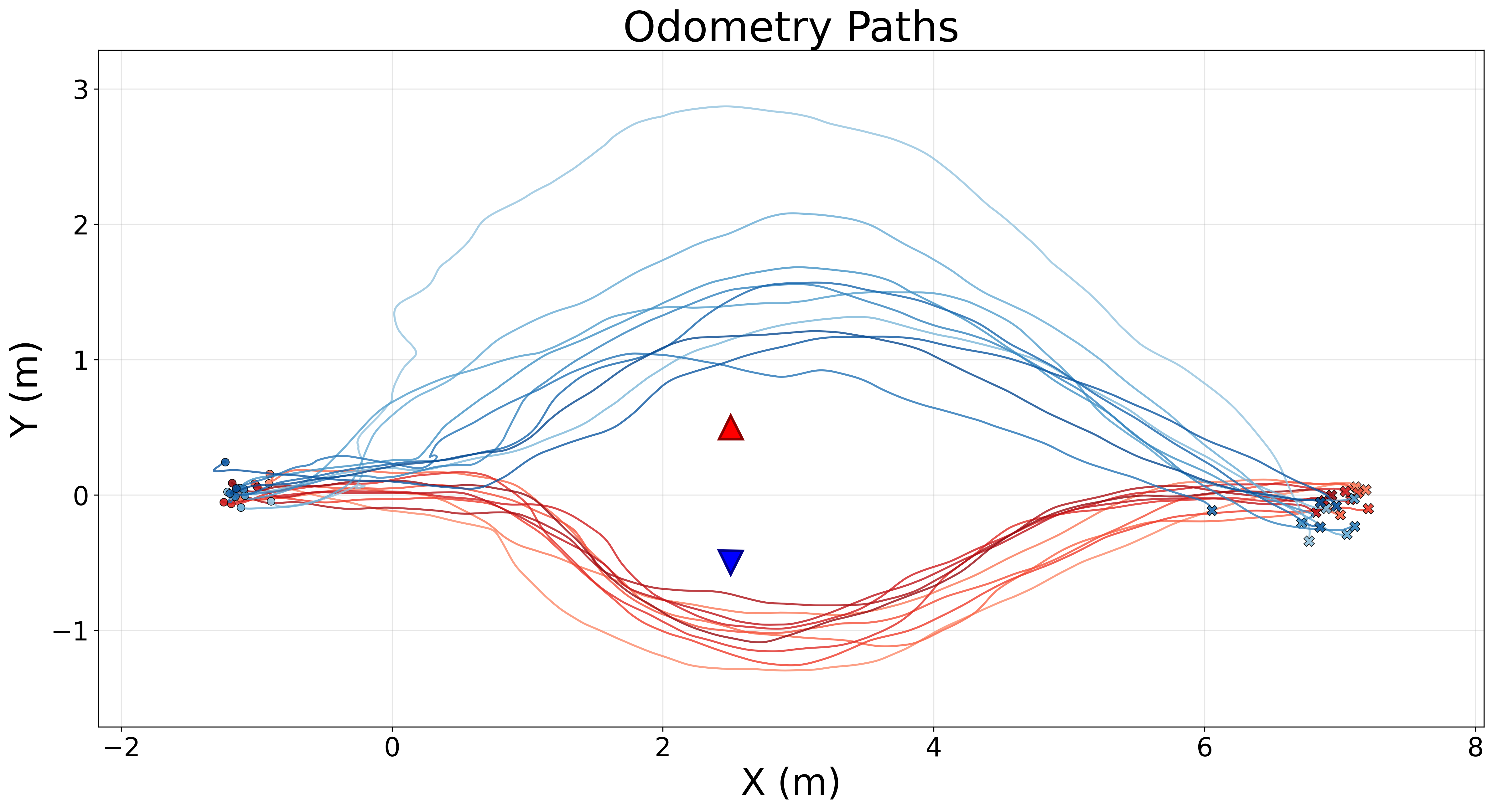}
  \end{subfigure}
  \begin{subfigure}{0.8\linewidth}
      \centering
      \includegraphics[width=\linewidth]{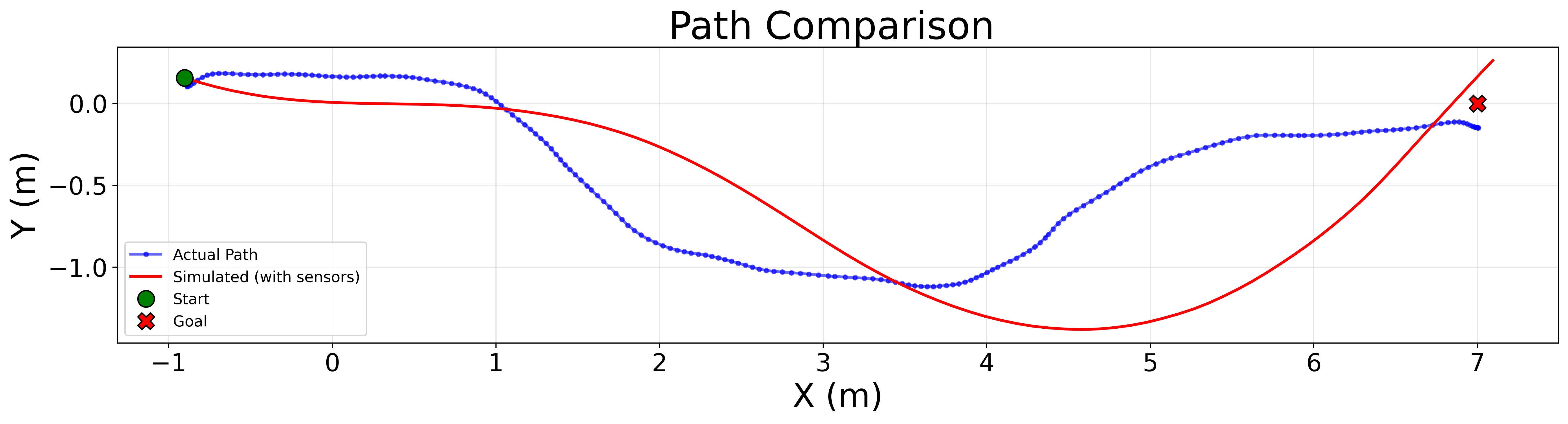}
  \end{subfigure}
  \caption{Top: Twenty paths recorded by a human operator. The blue paths were recorded with a fire at the blue position, requiring a left steering input and rightward correction after passing the fire. The red paths use an analogous setup in which the fire position is mirrored across the straight line. Bottom: One of the recorded paths (blue) compared to a path generated by inputting the original sensor reading into the learned policy (red). The fire is located at (2.5, 0.5).}
  \label{fig:human-paths-v2}
  \vspace{-4mm}
\end{figure}

\paragraph{Results}The operator was able to record 20 paths in under 30 minutes, generating a dataset that could be used to successfully train a small behavioral cloning model. The model notably steers later than the operator as it has to rely on the sensor data increasing substantially before committing to an avoidance direction and later correcting back to aim straight at the goal (see Fig.~\ref{fig:human-paths-v2}). 

\subsection{High-Frequency Thermally Reactive Control}
\label{sub:reactive-control}

This experiment demonstrates how virtual sensors measuring the incident thermal radiation can be used to reactively steer the robot away from a fire hazard.

\paragraph{Setup} The robot is augmented with four virtual thermal sensors at the corners of its bounding volume. One sensor each at the front-left, front-right, rear-right and rear-left corner, at the height of the robot's body. Each sensor $i$ measure the incoming thermal radiation $\dot{q_i}$. The sensors then output a velocity vector pointing towards the robot's geometric center and planar with the sensors. To steer the robot towards a goal, a constant velocity vector $v_t$ is added, that points from the robot to the goal. The sensors' velocities are scaled by a maximum radiation $\dot{q}_{max}$ to control the influence of the radiation on the steering. All of these velocities are aggregated:
\begin{equation}
    v' = \sum \frac{\dot{q_i}}{\dot{q}_{max}} + v_t,
\end{equation}
and then normalized to get the robot's final velocity:
\begin{equation}
    v = \frac{v'}{||v'||_2}.
\end{equation}
For the experiment environment, we used a simple plane on which we placed the robot, a single fire and the goal on a line. The full setup is shown in Fig.~\ref{fig:reactive-control-snapshots}.

\paragraph{Results} The robot starts in the vicinity of the fire and has to reach a goal on the other side of the fire. The temporal progression of the experiment is shown in Fig.~\ref{fig:reactive-control-snapshots}. The corresponding radiation sensor readings and odometry trajectory are shown in the plots in Fig.~\ref{fig:reactive-control-plots}. The robot starts a safe distance from the fire (a) without any thermal irradiance, as shown by the radiation sensors. The robot moves towards the goal behind the fire, however, as the radiation sensor readings increase at around 8s, the robot stops approaching the fire (b). The high radiation readings steer the robot around the fire (c, d) and also in the odometry trajectory. As the robot passes the fire, the left sensor readings decrease and the right sensor readings increase. After the robot has passed the fire at around 25s, the rear sensors detect more radiation than the front sensors, allowing the robot to move towards the goal unhindered. After the robot passed the fire, it steers straight towards the goal (e, f).

\subsection{Performance and Latency Evaluation}
To evaluate our asynchronous architecture against high-latency, we artificially increase the delay between the robot simulation and the combustion simulation on the thermally reactive control setup (see~\ref{sub:reactive-control}). Without any artificial delay, the simulation round-trip time was around 100 ms. When artificially adding a delay of up to 1 second, the robot was still able to avoid the fire and reach the goal. Only after adding at least 2 seconds of delay (high latency), the robot could not avoid the fire in time anymore, which led to a substantially increased heat exposure. This indicates that our framework is robust against significant variances in latency.

\section{CONCLUSIONS}

We present FaaS, a co-simulation framework that bridges the gap between the high-fidelity thermodynamic modeling required for fire-response robotics and the real-time, closed-loop demands of modern robot simulators. By treating fire as an asynchronous external service rather than a tightly coupled simulation component, FaaS delivers physically accurate thermal sensing and visually consistent flame and smoke rendering at  real-time rates, while remaining engine-agnostic across Isaac Sim, Gazebo, and MuJoCo.

Our experiments validate that the framework's thermodynamic outputs are spatially plausible for path planning
on standard ROS~2 costmaps, smooth for data collection and training closed-loop policies via Behavioral Cloning, and temporally responsive for safety-critical reactive controllers. 
A limitation of our approach is that the robot simulator has no impact on the fluid simulator beyond the initial shared scene setup. A robot walking into the fire therefore won't affect the fluid simulation. Furthermore, we do not simulate any damage done to the robot due to excessive heat exposure like material degradation or failing camera streams.

Future work will explore bidirectional coupling between fire dynamics and robot environments, including material-dependent combustion, structural degradation, and adaptive fire spread influenced by robotic actions. The current asynchronous architecture favors real-time robot control; a synchronous lockstep mode could additionally serve as an offline rendering option for applications prioritizing temporal accuracy over interactive rates. We also plan to extend the virtual sensor suite to include gas concentration and visibility estimation, supporting smoke-aware perception and planning.






\section*{ACKNOWLEDGMENTS}
This work has taken place in the VCAI Lab at Kiel University and in the RobotiXX Lab at George Mason University. The work is supported by ERC grant 101170158 - WildfireTwins. RobotiXX research is supported by National Science Foundation (NSF, 2350352), Army Research Office (ARO, W911NF2320004, W911NF2520011), Google DeepMind (GDM), Clearpath Robotics, FrodoBots Lab, Raytheon Technologies (RTX), Tangenta, Mason Innovation Exchange (MIX), and Walmart. We also acknowledge technical support by the Embodied AI Center at Kiel University.


\bibliographystyle{IEEEtran}
\bibliography{references}

\end{document}